\documentclass{article}

\PassOptionsToPackage{square,numbers}{natbib}

\usepackage[preprint]{neurips_2025}
\usepackage{graphicx}
\usepackage{amsmath, amssymb, amsthm}
\usepackage{algpseudocode}
\usepackage{algorithm}
\usepackage{appendix}
\usepackage{multirow}
\usepackage{booktabs}
\usepackage{subfigure}

\usepackage[utf8]{inputenc} 
\usepackage[T1]{fontenc}    
\usepackage{hyperref}       
\usepackage{url}            
\usepackage{booktabs}       
\usepackage{amsfonts}       
\usepackage{nicefrac}       
\usepackage{microtype}      
\usepackage{xcolor}         

\newtheorem{theorem}{Theorem}[section]
\newtheorem{lemma}{Lemma}[section]
\title{Gradual Binary Search and Dimension Expansion : A general method for activation quantization in LLMs}

\author{%
  Lucas Maisonnave  \\
  Université Paris-Saclay \\
  CEA, List
  \\F-91120, Palaiseau, France \\
  \texttt{lucas.maisonnave@cea.fr} \\
   \And
   Cyril Moineau \\
   Université Paris-Saclay \\
  CEA, List
  \\F-91120, Palaiseau, France \\
  \texttt{cyril.moineau@cea.fr} \\
  \And
   Olivier Bichler \\
   Université Paris-Saclay \\
  CEA, List
  \\F-91120, Palaiseau, France \\
  \texttt{olivier.bichler@cea.fr} \\
  \And
  Fabrice Rastello \\
  Univ. Grenoble Alpes, \\
  Inria, CNRS, Grenoble INP, LIG, \\
  38000 Grenoble, France \\
\texttt{fabrice.rastello@inria.fr} \\
}

\begin{document}

\maketitle

\begin{abstract}

Large language models (LLMs) have become pivotal in artificial intelligence, demonstrating strong capabilities in reasoning, understanding, and generating data. However, their deployment on edge devices is hindered by their substantial size, often reaching several billion parameters. Quantization is a widely used method to reduce memory usage and inference time, however LLMs present unique challenges due to the prevalence of outliers in their activations. In this work, we leverage the theoretical advantages of Hadamard matrices over random rotation matrices to push the boundaries of quantization in LLMs. We demonstrate that Hadamard matrices are more effective in reducing outliers, which are a significant obstacle in achieving low-bit quantization. Our method based on a gradual binary search enables 3-bit quantization for weights, activations, and key-value (KV) caches, resulting in a 40\% increase in accuracy on common benchmarks compared to SoTA methods. We extend the use of rotation matrices to support non-power-of-2 embedding dimensions, similar to the Qwen architecture, by employing the Paley's algorithm. Our experimental results on multiple models family like Mistral, LLaMA, and Qwen demonstrate the effectiveness of our approach, outperforming existing methods and enabling practical 3-bit quantization.
\end{abstract}

\section{Introduction}

Large Language Models (LLMs) have become a central component of artificial intelligence due to their strong capabilities in reasoning, understanding, and generating data. These impressive capabilities are attributed to the quality of the data used during training, the model architecture, and the size of the model, which often reaches several billion parameters. This size limitation restricts their deployment on edge devices. Quantization is a widely used method to reduce memory usage and inference time \cite{gholami_survey_2021,guo_survey_2018}, but the challenges differ compared to those faced with Convolutional Neural Networks (CNNs) \cite{esser_learned_2020,xiao_smoothquant_2023}. 

Weights are relatively easy to quantize for both CNNs and LLMs and can often achieve ternary quantization without significant loss of accuracy \cite{ma_era_2024,zhu_trained_2017}. However, activations behave differently in transformer architectures \cite{nrusimha_mitigating_2024}. The presence of outliers in activations makes conventional quantization (symmetric uniform) very challenging, hindering our ability to achieve 4-bit quantization. LLMs are known to produce spikes in its layers and for some tokens that can be handled separately or diffused in the tensor \cite{dettmers_llmint8_nodate,xiao_smoothquant_2023}.

One very promising approach to overcome this limitation is to use rotation matrices to redistribute weights and activation values, thereby minimizing the impact of outliers \cite{liu_spinquant_2024,ashkboos_quarot_2024}. Additionally, methods such as prefix tokens have shown very interesting results in managing outliers in LLMs \cite{chen_prefixquant_2024,son_prefixing_2024}.

In this work, we leverage results on rotation matrices to push the boundaries further and enable 3-bit Weights, Activations, KV cache (WAKV) quantization by employing a binary search. We extend this method to a more general approach capable of handling non-power-of-2 embedding dimensions, similar to Qwen. Our main contributions are:

\begin{itemize}
\item A theoretical demonstration that Hadamard matrices are more effective in reducing outliers than rotation matrices drawn on the unit sphere.
\item 3-bit quantization for weights, activations, and KV cache, resulting in a 40\% increase in accuracy on common benchmarks using a gradual binary search.
\item Extension of rotation matrices to support non-power-of-2 embedding dimensions using the Paley's algorithm.
\item The introduction of dimension expansion to build a more general rotation pipeline allowing architectures like Qwen to work with rotations.

\end{itemize}

\section{Related Works}

\subsection{Quantization}

Quantizing models involves reducing the number of bits required to store and compute model activations. This process is crucial for deploying LLMs on resource-constrained devices. To achieve this, we define a scaling factor that determines the distance between quantization bins and the range of values to be compressed.

For symmetric uniform quantization, we apply a rounding function to a scaled distribution:
$$\hat X = \text{round}\left( \frac{X}{\Delta}\right )\Delta,\, \Delta= \frac{\max |X|}{2^b - 1}$$
 
where $\Delta$ is the scaling factor, $b$ is the bitwidth, and $\max |X|$ is the maximum absolute value of the distribution, preserving extreme values for activations.

Such quantization can be applied per-token, where each token has a different scaling factor, or per-tensor, where a single scaling factor is used for each activation tensor \cite{gholami_survey_2021,guo_survey_2018}. Per-token quantization is more challenging to implement efficiently in practice compared to per-tensor quantization but results in better quantization performances. Scaling factors can be static during inference, based on statistics computed on a subset of the dataset, or dynamic, recomputed at each step.

Quantization can lead to a significant drop in performance when applied post-training (PTQ) \cite{yang_post-training_2023}. To mitigate this, some methods adapt weights to the noise introduced during a training phase (QAT) \cite{lin_defensive_2019,defossez_differentiable_2022}. Typically, for LLMs, only linear layers are quantized, as they account for most of the computation cost, while normalization layers, matrix multiplications, and the softmax function within the attention block are left unquantized.

\subsection{Outliers}

Quantizing LLM weights is relatively straightforward and does not require extensive efforts to achieve. Techniques like GPTQ \cite{frantar_gptq_2023} enables 8-bit quantization without retraining, preserving model accuracy. Some QAT methods can even push the boundaries to 1-bit quantization, as seen in BitNet \cite{wang_bitnet_2023} or ternary quantization \cite{ma_era_2024}.

However, LLMs present unique challenges due to the prevalence of extreme high values in their activations \cite{wei_outlier_2023,nrusimha_mitigating_2024,huang_rolora_2024,lin_duquant_nodate}. The scaling factor, which is directly tied to the maximum absolute value, often causes most of the distribution to be rounded to zero, leading to performance degradation. To address this, techniques like LLM.int8() \cite{dettmers_llmint8_nodate} cluster these outliers and quantize them separately from the main distribution.

Alternative methods, such as SmoothQuant \cite{xiao_smoothquant_2023}, shift the quantization challenge from activations to weights by introducing a scaling parameter between them. Other approaches attempt to relocate these spikes into "sink tokens" before quantization \cite{son_prefixing_2024}. Some research focuses on understanding the upstream causes of these spikes during the learning process to limit their impact post-training \cite{nrusimha_mitigating_2024}. Additionally, efforts are made to better locate these outliers by visualizing the layers, dimensions, and tokens that may be their source \cite{maisonnave_precision_2025}.

\begin{figure}[t]
    \centering
    \includegraphics[width=1\textwidth]{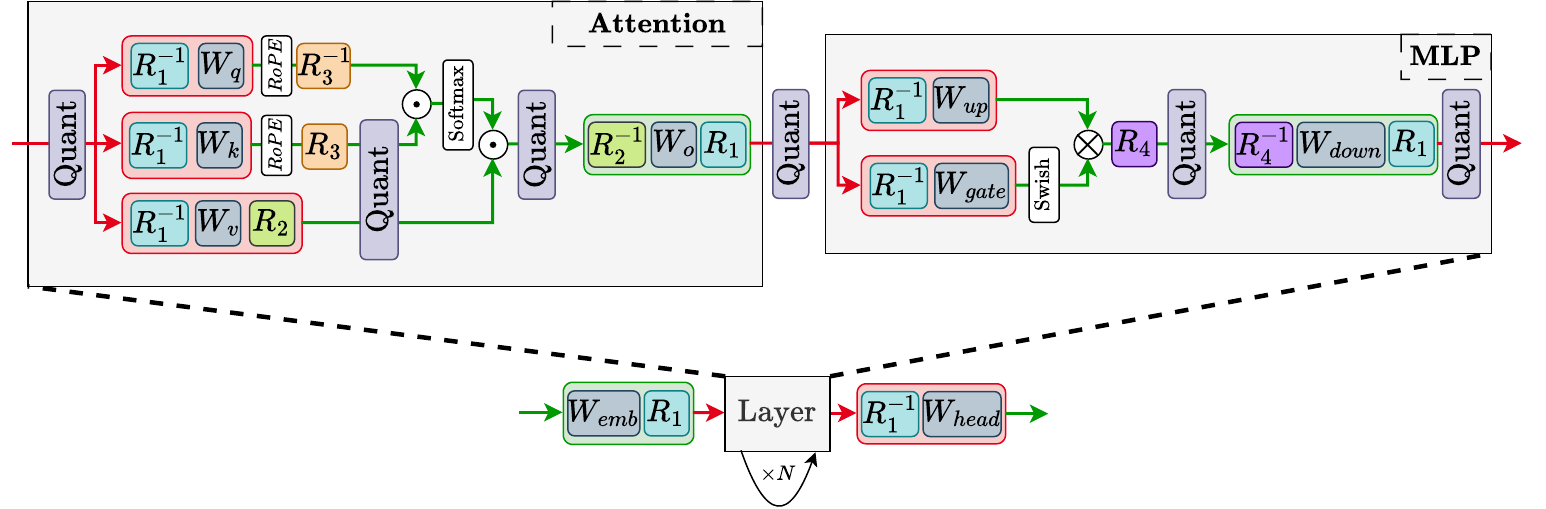}
    \caption{Architecture's pipeline with rotation matrices $R_1, \, R_2,\, R_3,\, R_4$ and dimension expansion. Red lines represents expanded tokens in $4096 + d$ dimensions and green lines represents non expanded tokens. Projections in red (QKV, Gate, Up and $LM_{head}$) have their input weigths dimension expanded and projections in green (Out, Down and Embeddings) have their output weights dimension expanded} 
    \label{fig:pipeline_expanded}
\end{figure}

\subsection{Rotation Matrices}

\subsubsection{Random orthogonal matrices}

Rotation matrices play a pivotal role in various applications, including signal processing, computer vision, and machine learning. These matrices are orthogonal and invertible by their transpose, meaning they preserve the length of vectors and the angles between them. In the context of quantization, rotation matrices can be used to decorrelate and redistribute the energy of model activations \cite{ashkboos_slicegpt_2024,ashkboos_quarot_2024,chee_quip_nodate}, making them more amenable to quantization. The idea is to apply orthogonal matrices before quantization to flatten the distribution and then recover the tensor by applying its inverse (see Figure \ref{fig:pipeline_expanded}). Part of this process can be pre-computed and fused with weights and the rest needs to be done at inference \cite{ashkboos_slicegpt_2024}.

However, the effectiveness of rotation matrices depends on the specific matrix used. Randomly drawn orthogonal rotation matrices can introduce noise and reduce the overall performance of the model. To mitigate this, some methods adapt the rotation matrices during a training phase to better align with the model's weights and activations \cite{liu_spinquant_2024}.

In practice, rotation matrices are often used in conjunction with other quantization techniques, such as GPTQ. This combination allows for more robust and efficient quantization of large language models, enabling their deployment on resource-constrained devices.

\subsubsection{Hadamard matrices}
\label{sec:hadamard_matrix}
Hadamard matrices are another powerful tool in the quantization arsenal. These matrices are orthogonal matrices and all their entries are either +1 or -1 making them very efficient to compute (eq \ref{eq:hadamard}). Hadamard matrices have been extensively used in signal processing, error-correcting codes \cite{horadam_hadamard_2012}, and more recently, in the quantization of neural networks \cite{ashkboos_slicegpt_2024,ashkboos_quarot_2024}.

One of the key advantages of Hadamard matrices is their ability to decorrelate the activations of a model. By applying a Hadamard matrix, the activations are transformed into a new basis where the correlations between different dimensions are minimized. This decorrelation property is particularly useful in reducing the impact of outliers, as the extreme values are spread out across multiple dimensions.

Hadamard matrices of order $ 2^n $ can be constructed recursively using the Fast Hadamard Transform (FHT) method: For $ n \geq 1 $, construct the $ 2^{n+1} \times 2^{n+1} $ Hadamard matrix $ H_{2^{n+1}} $ using the $ 2^n \times 2^n $ Hadamard matrix $ H_{2^n} $ as follows:
\begin{equation}
\label{eq:hadamard}
H_2 = \begin{pmatrix}
1 & 1 \\
1 & -1
\end{pmatrix}
, \quad
H_4 = \begin{pmatrix}
1 & 1 & 1 & 1 \\
1 & -1 & 1 & -1 \\
1 & 1 & -1 & -1 \\
1 & -1 & -1 & 1 
\end{pmatrix}, \quad
H_{2^{n+1}} = \begin{pmatrix}
H_{2^n} & H_{2^n} \\
H_{2^n} & -H_{2^n}
\end{pmatrix}
\end{equation}

This method is highly efficient for generating Hadamard matrices and can be applied in real-time. In summary, both rotation matrices and Hadamard matrices are essential for the quantization of large language models. However, Hadamard matrices offer several advantages: they can be generated more efficiently, their structure of containing only 1 and -1 makes them highly efficient for matrix multiplication, and they are known to handle outliers in activations more effectively \cite{liu_spinquant_2024}. In the following sections, we will theoretically demonstrate that Hadamard matrices are more effective than random rotation matrices drawn from the unit sphere in reducing the amplitude of outliers.

\subsubsection{Paley algorithm}

To generate other dimensions $n$ for Hadamard matrix we can use known small matrices and apply power of 2 algorithm as used in QuaRot \cite{ashkboos_quarot_2024} but it can be limiting and doesn't cover a lot of values. To overcome this issue we can use the Paley's Algorithm that generate a Hadamard matrix $n\times n$ if $n - 1$ is a prime number and $n-1 \equiv 3 \pmod 4 $. This algorithm is described below (Algorithm  \ref{alg:paley}) and needs to generate Legendre symbols \(\left(\frac{a}{p}\right)\) which take any integer number $a$ and prime number $p$ to produce a value in $\{-1,0,1\}$ as below :
\begin{itemize}
    \item If \( a \) is a quadratic residue modulo \( p \), then there exists an integer \( x \) such that \( x^2 \equiv a \pmod{p} \). In this case, \(\left(\frac{a}{p}\right) = 1\).
    
    \item If \( a \) is a quadratic non-residue modulo \( p \), then there is no integer \( x \) such that \( x^2 \equiv a \pmod{p} \). In this case, \(\left(\frac{a}{p}\right) = -1\).
    
    \item If \( a \equiv 0 \pmod{p} \), then \(\left(\frac{a}{p}\right) = 0\).
\end{itemize}

Generating Legendre symbols can be time-consuming, especially for high-dimensional matrices. However, in the following sections, we will use this algorithm to generate non-power-of-2 Hadamard matrices and fuse them with the weights, so we only need to compute the Legendre symbols once.

\section{Analysis and theoretical demonstrations}
\subsection{Clipping Ratio}
\label{subsec:clip_ratio}

To perform quantization we can play on several parameters to improve the effectiveness of the process, for example in LSQ \cite{esser_learned_2020} they optimise the scaling factor trough training, or FracBits \cite{yang_fracbits_2021} which tries to find the best precision for every layer. Other works highlighted the importance of the clipping ratio like PACT \cite{choi_pact_2018} where the optimization is done during training. Some others apply a Grid Search \cite{chen_prefixquant_2024} to find the best configuration particularly useful for LLMs where training or fine tuning can be very time consuming.

Clipping ratios are essential for managing outliers, as they establish the balance between maintaining high precision for small values and preserving a maximum value close to its original. However, the model exhibits significant variability in how quantization responds to changes in the clipping ratio. While some projections can tolerate very low clipping ratios, others experience a substantial accuracy drop with even slight adjustments (see Appendix \ref{app:GBS_process}). Therefore, to effectively manage this variability, a tailored clipping ratio must be determined for each projection.

Previous studies have shown that quantization error is not always the best metric to guide the optimization process for quantization parameters \cite{maisonnave_applying_nodate}. Specifically, at very low precision levels, such as 4 or 3 bits, the set of quantized weights deviates significantly from the optimized configuration obtained during training. Attempting to recover this configuration using quantization error often results in an ineffective set of weights. To address this issue, we can use perplexity as an objective function. Perplexity provides a more accurate representation of model performance and is computationally efficient, as it is based on Cross Entropy Loss, which is frequently used during training for its smoothness.

\subsection{Hadamard Matrices reduce outliers more}
\label{sec:hadamard}
Experimentally, it is observed that Hadamard matrices tend to reduce better the amplitude of outliers present in the layers of LLMs, which directly impacts the performance of these models. However, the question of why such a phenomenon occurs has remained open from a theoretical perspective. We now provide an answer to this question.

\begin{theorem}[Hadamard reduction]
$\forall x \in \mathbb{R}^n$ containing an outlier, i.e., $x = (c, \epsilon, ..., \epsilon)^T$ with $c >> \epsilon$ we have
$$\max_{1 \leq i \leq n}|(Hx)_i| \leq \max_{1 \leq i \leq n}|(Qx)_i|$$
with $H$ a Hadamard matrix belonging to $\mathbb{R}^{n \times n}$ and $Q$ a rotation matrix drawn randomly on the unit sphere $\mathcal{S}^{n-1} = \{x \in \mathbb{R}^n: ||x||_2 = 1\}$.
\label{thm:HdmRed}
\end{theorem}

To demonstrate Theorem \ref{thm:HdmRed}, we can calculate the two terms of the inequality and thus show the superiority of one over the other.

\begin{lemma}[Hadamard incoherence] For $H$ a Hadamard matrix belonging to $\mathbb{R}^{n \times n}$ and $x = (c, \epsilon, ..., \epsilon)^T$ with $c >> \epsilon$
$$\max_{1 \leq i \leq n}|(Hx)_i| = \frac{c}{\sqrt{n}}$$
\label{lemme:1}
\end{lemma}

\begin{lemma}[Rotation incoherence] For $Q$ a rotation matrix drawn randomly on the unit sphere $\mathcal{S}^{n-1} = \{x \in \mathbb{R}^n: ||x||_2 = 1\}$ and $x = (c, 0, ..., 0)^T$ with $c >> 1$
$$\max_{1 \leq i \leq n}|(Qx)_i| = c\sqrt{\frac{2\log n}{n}}$$
\label{lemme:2}
\end{lemma}

\begin{algorithm}[t]
\caption{Gradual Binary Search}
\footnotesize     
\label{alg:grid_search}
\begin{algorithmic}[1]
\Require A model \( M \), a dataset $D$, a threshold $\epsilon$
\Ensure A list L of clipping ratios \Comment{+ operand on L means concatenation}
\State \( n \gets \text{number of projections in M} \)
\State $L \gets [\,]$
\For{\( i \gets 1 \) to \( n-1 \)}
    \State \(a \gets 0\)
    \State \(b \gets 1\)
    \State \(m \gets (a + b) / 2\) \Comment{We keep track of the middle element}
    \State \(M \gets \text{quantize\_proj}(M, i) \) \Comment{Quantize projection $i$ of model $M$}
    \State \(f_m = \text{evaluate}(M, D, L + m)\) \Comment{Evaluate model $M$ on dataset $D$ with clipping ratios L}
    \State \(\text{iteration} \gets 0 \)
    \While{$b - a > \epsilon$} \Comment{We iterate until we converge}
        \If{$\text{iteration is even}$} \Comment{Allows to use only one loop for binary search}
            \State $x \gets (a + m)/2$
        \Else
            \State $x \gets (b + m)/2$
        \EndIf
        \State $f_x \gets \text{evaluate}(M, D, L + x)$ \Comment{Evaluate model on a new clipping ratio}
        \If{$f_x < f_m$} \Comment{If we improve PPL (the lower the better) we keep it}
            \If{$x < m$} \Comment{If the target is less than the middle, search the left half}
                \State $b \gets m$
            \Else \Comment{If not , search the right half}
                \State $a \gets m$
            \EndIf
            \State $m, f_m \gets x, f_x$
        \Else
            \If{$x < m$}
                \State $a \gets x$
            \Else
                \State $b \gets x$
            \EndIf
        \EndIf
        \State $\text{iteration} \gets \text{iteration} + 1$
    \EndWhile
    \State $L = L + m$ \Comment{Add new element to the list}
\EndFor
\State \Return \( L \)
\end{algorithmic}
\end{algorithm}

We can prove in Lemma \ref{lemme:1} and Lemma \ref{lemme:2} that the reduction of outliers with a Hadamard matrix is of order $O(\frac{1}{\sqrt{n}})$ and $O(\sqrt{\frac{2\log n}{n}})$ for a random orthogonal matrix (demonstrations are done in Appendix \ref{app:theoretical_results}). These results prove Theorem \ref{thm:HdmRed} and also show the close link between reduction and the dimension of embeddings in LLMs. The higher the dimension is the stronger the reduction will be. We can also demonsatrate in Theorem \ref{thm:HdmOpt} that Hadamard matrices are optimal and the best group of matrices to reduce outliers.

\begin{theorem}[Hadamard optimality]
For any orthogonal matrix $Q \in \mathbb{R}^{n\times n}$ we have:
$$
\max_{1\leq i,j \leq n}|Q_{ij}| \geq \frac{1}{\sqrt{n}}
$$
\noindent
And a Hadamard matrix reaches this bound.
\label{thm:HdmOpt}
\end{theorem}

\section{Method}
\subsection{Gradual Binary Search}

In Section \ref{subsec:clip_ratio}, we emphasize the importance of the clipping ratio parameter and its significant impact on model performance. We stress the need to optimize each projection with its own clipping ratio for best results. Our primary contribution is an algorithm that determines the optimal clipping ratio for each quantizer using a binary search (Algorithm \ref{alg:grid_search}). To drive the binary search, we minimize perplexity across various clipping ratios, assuming a single minimum and a convex landscape. Additionally, we quantize our model gradually: first, we quantize and optimize the initial linear projection while keeping the rest in FP16, then use the obtained parameters to quantize and optimize the next projection, and so on. This process is discussed in Appendix \ref{app:GBS_process} where we experimentally show the necessity to optimize gradually the clipping ratio.

\subsection{Increasing dimensions}
\label{sec:incr_dim}
\begin{lemma}[Expanding limit] For a matrix product $AB$ with $A \in \mathbb{R}^{m\times n}$ and $B \in \mathbb{R}^{n\times p}$ in $b$ bits and $A'B'$ with $A' \in \mathbb{R}^{m\times (n+d)}$ and $B' \in \mathbb{R}^{(n+d)\times p}$ in $b'$ bits we must have $d \leq \frac{n(b-b')}{b'}$ so that $\text{BitOps}(A'B') \leq \text{BitOps}(AB)$, with $m, \, n, \, p , \, b , \, b' \in \mathbb{N}$ and $b' \leq b$
\label{lemme:3}
\end{lemma}

One important limitation of QuaRot's implementation of rotation matrices in LLMs is the necessity to have embeddings in a power of 2 dimension which can be very limiting in some architectures like Qwen2.5-1.5B which have features in dimension 1536 and can not be quantized with QuaRot. To overcome this problem we increase manually the dimension of embedding by adding zeros in the weights (independently developed in \cite{franco_improving_2025}) to reach a dimension suitable to generate a Hadamard matrix with the Paley's algorithm \ref{alg:paley}. Then we save the matrix product of weights padded with 0s and the Hadamard matrix as our new weights (see figure \ref{fig:pipeline_expanded} and Eq \ref{eq:expand} for an example in dimension 4). The primary goal is to create a more versatile pipeline compatible with any architecture but it also enhance performance through increased dimensionality.
\definecolor{myblue}{RGB}{0, 0, 255}
\definecolor{myred}{RGB}{255, 0, 0}
\begin{equation}
    W \gets \begin{pmatrix}
\textcolor{myblue}{1 }& \textcolor{myblue}{1 }& \textcolor{myred}{1 }& \textcolor{myred}{1 }\\
\textcolor{myblue}{1 }& \textcolor{myblue}{-1} & \textcolor{myred}{1 }& \textcolor{myred}{-1} \\
\textcolor{myred}{1 }& \textcolor{myred}{1 } & \textcolor{myred}{-1} & \textcolor{myred}{-1}\\
\textcolor{myred}{1 }& \textcolor{myred}{-1} & \textcolor{myred}{-1} & \textcolor{myred}{1
}\end{pmatrix} \times \begin{pmatrix}
\textcolor{myblue}{a} & \textcolor{myblue}{b} \\
\textcolor{myblue}{c} & \textcolor{myblue}{d} \\
\textcolor{myred}{0} & \textcolor{myred}{0} \\
\textcolor{myred}{0} & \textcolor{myred}{0}
\end{pmatrix}
\label{eq:expand}
\end{equation}

Indeed theorem \ref{thm:HdmRed} ensures that increasing the dimension helps reduce the impact of outliers in any tensor. Consequently, by adding zeros to the weight tensors, we also improve the effectiveness of quantization. The intuition behind it is that by adding more dimensions in our tensors we create more space to store information and especially outliers which will be sliced in more parts and recovered better after quantization. This process increases the model size and computational cost, necessitating a trade-off to achieve better accuracy without a significant increase in computational requirements.

Since we only expand the input and output dimensions of the attention block and Multi-Layer Perceptron (MLP), there is no additional computational cost at inference. The new weights are stored, and the Hadamard matrices required for inference remain unchanged, allowing them to be efficiently computed using the FHT.

Lemma \ref{lemme:3} shows the threshold after which the increase in dimentionality is worse than just quantizing with one more bit. For example with a LLaMA3-8B which has embeddings in 4096 dimensions we are only allowed to increase to $d=1366$ dimensions in 3 bits before reaching the computational cost in 4 bits.

\section{Experiments}
\subsection{Setup}

We conduct our experiment based on the the code of QuaRot which performs per-token quantization for activations and GPTQ for weights. We also quantize KV caches using asymmetric quantization with a group size of 128. We compare our results on several metrics : perplexity (PPL) on WikiText2, and 6 benchmarks : PIQA, hellaswag (HS), arc-easy (ARC-E), arc-challenge (ARC-C), winogrande (WINO) and lambada, we also compute the average value (AVG) of these 6 benchmarks. We performs ours experiments in 4 and 3 bits quantization on 6 different models from the Mistral library, LLaMA architecture and Qwen. We used only one GPU A100 to perform quantization and Gradual Binary Search (GBS) with 10\% of the train set of WikiText2 for 4 days for the biggest models.

\begin{center}
    
\begin{table}[t]
    \footnotesize
    \renewcommand{\arraystretch}{1.05}
    \setlength{\tabcolsep}{3pt}
    \centering
    \caption{Results in 4 bits WAKV quantization on perplexity (PPL), PIQA, hellaswag (HS), arc-easy (ARC-E), arc-challenge (ARC-C), winogrande (WINO) and lambada, we also compute the average value (AVG) which represents a \% of success. We compare our method, GBS, with QuaRot (where $\text{QuaRot}^{+}$ indicates the use of dimension expansion) and clearly observe that GBS outperforms QuaRot across all computed metrics. Additionally, dimension expansion enables QuaRot to be compatible with Qwen's family of models.}
    \begin{tabular}{cccccccccc}
        \toprule
        \textbf{Model} & \textbf{Method} & \textbf{PPL}$\downarrow$ & \textbf{PIQA} & \textbf{HS} & \textbf{ARC-E} & \textbf{ARC-C} & \textbf{Wino} & \textbf{Lambada} & \textbf{AVG}$\uparrow$ \\
        \hline
        \multirow{3}{*}{Mistral 7B Inst v0.3} & FP16 & 5.49  & 71.27 & 74.6 & 67.94 & 74.27 & 58.96 & 82.66 & 71.62 \\
                                                  & QuaRot & 5.98  & 67.7 & 71.45 & 63.94 & 69.53 & 55.2 & 79.46 & 67.88 \\
                                                  & QuaRot + GBS & \textbf{5.75}   & \textbf{70.55} & \textbf{74.44} & \textbf{66.66} & \textbf{71.82} & \textbf{56.66} & \textbf{80.77} & \textbf{70.15} \\
        \hline
        \multirow{3}{*}{Mistral 7B v0.1} & FP16 & 5.25 & 72.49 & 75.59 & 69.4 & 73.95 & 54.86 & 80.18 & 71.08 \\
                                                  
                                                  & QuaRot & 5.82 & 67.62 & 71.88 & 63.36 & 70.01 & 48.98 & 76.6 & 66.41 \\
                                                  & QuaRot + GBS & \textbf{5.57} & \textbf{71.47} & \textbf{74.95} & \textbf{68} & \textbf{72.22} & \textbf{51.54} & \textbf{79.21} & \textbf{69.56} \\
        \hline
        \multirow{3}{*}{Llama2 7B} & FP16 & 5.47 & 71.08 & 73.9 & 68.25 & 68.98 & 46.33 & 74.58 & 67.19 \\
                                                  
                                                  & QuaRot & 6.21 & 64.65 & 69.09 & 60.22 & 64.64 & \textbf{43.17} & 69.78 & 61.92 \\
                                                  & QuaRot + GBS & \textbf{6.04} & \textbf{65.64} & \textbf{69.88} & \textbf{61.4} & \textbf{66.46} & 42.32 & \textbf{70.75} & \textbf{62.74} \\
        \hline
        \multirow{3}{*}{Llama3 8B} & FP16 & 6.13 & 72.62 & 76.01 & 69.22 & 72.93 & 53.41 & 77.69 & 70.3 \\
                                                  
                                                  & QuaRot & 8.33 & 61.66 & 66.27 & 57.05 & 64.72 & 42.06 & 68.06 & 59.97 \\
                                                  & QuaRot + GBS & \textbf{7.4} & \textbf{67.87} & \textbf{72.02} & \textbf{63.73} & \textbf{71.03} & \textbf{45.9} & \textbf{73.7} & \textbf{65.71} \\
        \hline
        \multirow{3}{*}{Qwen2.5 7B Inst} & FP16 & 7.45 & 66.23 & 69.73 & 62.74 & 70.56 & 55.12 & 81.02 & 67.57 \\
                                                  & QuaRot & - & - & - & - & - & - & - & - \\
                                                  & $\text{QuaRot}^{+}$ & 9.21 & 56.66 & 59.31 & 54.01 & 63.54 & 48.89 & 69.78 & 58.7 \\
                                                  & $\text{QuaRot}^{+}$ + GBS & \textbf{8.23} & \textbf{62.58}  & \textbf{64.91}  & \textbf{60.24}  & \textbf{66.61}  & \textbf{49.4}  & \textbf{72.39}  & \textbf{62.69} \\
        \hline
        \multirow{3}{*}{Qwen2.5 1.5B Inst} & FP16 & 9.64 & 58.09 & 61.21 & 54.98 & 63.3 & 46.59 & 75.8 & 60.0 \\
                                                 & QuaRot & - & - & - & - & - & - & - & - \\ 
                                                  & $\text{QuaRot}^{+}$ & 14.44 & 39.05 & 40.23 & 37.86 & 54.85& 35.75& 58.71 & 44.41 \\
                                                  & $\text{QuaRot}^{+}$ + GBS & \textbf{12.05} & \textbf{43.94} & \textbf{45.24} & \textbf{42.64} & \textbf{58.64} & \textbf{39.33} & \textbf{65.61} & \textbf{49.23} \\
        \bottomrule
        
    \end{tabular}
    
    \label{tab:results_4bits}
\end{table}
\end{center}

\begin{center}
\begin{table}[t]
    \footnotesize
    \renewcommand{\arraystretch}{1}
    \setlength{\tabcolsep}{3pt}
    \centering
    \caption{Results in 3 bits WAKV quantization on perplexity (PPL), PIQA, hellaswag (HS), arc-easy (ARC-E), arc-challenge (ARC-C), winogrande (WINO) and lambada, we also compute the average value (AVG) which represents a \% of success. We compare our method, GBS, with QuaRot (where $\text{QuaRot}^{+}$ indicates the use of dimension expansion) and clearly observe that GBS outperforms QuaRot across all computed metrics. Additionally, dimension expansion enables QuaRot to be compatible with Qwen's family of models.}
    \begin{tabular}{cccccccccc}
        \toprule
        \textbf{Model} & \textbf{Method} & \textbf{PPL}$\downarrow$ & \textbf{PIQA} & \textbf{HS} & \textbf{ARC-E} & \textbf{ARC-C} & \textbf{Wino} & \textbf{Lambada} & \textbf{AVG}$\uparrow$ \\
        \hline
        \multirow{3}{*}{Mistral 7B Inst v0.3} & FP16 & 5.49 & 71.27 & 74.6 & 67.94 & 74.27 & 58.96 & 82.66 & 71.62 \\
                                                  
                                                  & QuaRot & 38.28 & 7.66 & 9.49 & 5.82 & 51.46 & 23.55 &  34.39 & 22.06 \\
                                                  & QuaRot + GBS & \textbf{7.04} & \textbf{62.17} & \textbf{66.93} & \textbf{57.4} & \textbf{61.56} & \textbf{46.42} & \textbf{73.44} & \textbf{61.32} \\
        \hline
        \multirow{3}{*}{Mistral 7B v0.1} & FP16 & 5.25 & 72.49 & 75.59 & 69.4 & 73.95 & 54.86 & 80.18 & 71.08 \\
                                                  
                                                  & QuaRot & 100.85 & 3.07 & 4.13 & 2.0 & 48.62 & 22.18 & 30.6 & 18.43 \\
                                                  & QuaRot + GBS & \textbf{7.31} & \textbf{59.22} & \textbf{64.41} & \textbf{54.03} & \textbf{63.3} & \textbf{40.53} & \textbf{68.39} & \textbf{58.31} \\
        \hline
        \multirow{3}{*}{Llama2 7B} & FP16 & 5.47 & 71.08 & 73.9 & 68.25 & 68.98 & 46.33 & 74.58 & 67.19 \\
                                                  
                                                  & QuaRot & 332.56 & 0.25 & 0.47 & 0.04 & 51.14 & 26.11 & 30.39 & 18.07 \\
                                                  & QuaRot + GBS & \textbf{9.18} & \textbf{39.03} & \textbf{49.91} & \textbf{28.14} & \textbf{56.12} & \textbf{31.91} & \textbf{54.92} & \textbf{43.34} \\
        \hline
        \multirow{3}{*}{Llama3 8B} & FP16 & 6.13 & 72.62 & 76.01 & 69.22 & 72.93 & 53.41 & 77.69 & 70.3 \\
                                                  
                                                  & QuaRot & 1315 & 0.05 & 0.08 & 0.02 & 49.41 & 23.72 & 27.78 & 16.84 \\
                                                  & QuaRot + GBS & \textbf{12.62} & \textbf{44.92} & \textbf{50.32} & \textbf{39.51} & \textbf{60.22} & \textbf{32.85} & \textbf{53.7} & \textbf{46.92} \\
        \hline
        \multirow{3}{*}{Qwen2.5 7B Inst} & FP16 & 7.45 & 66.23 & 69.73 & 62.74 & 70.56 & 55.12 & 81.02 & 67.57 \\
                                                  & QuaRot & - & - & - & - & - & - & - & - \\
                                                  
                                                  & $\text{QuaRot}^{+}$ & 251 & 1.14 & 1.14 & 1.14 & 49.33 & 25.0 & 32.15 & 18.32 \\
                                                  & $\text{QuaRot}^{+}$ + GBS & \textbf{12.33} & \textbf{41.9} & \textbf{42.62} & \textbf{41.18} & \textbf{56.59} & \textbf{40.53} & \textbf{61.83} & \textbf{47.44} \\
        \hline
        \multirow{3}{*}{Qwen2.5 1.5B Inst} & FP16 & 9.64 & 58.09 & 61.21 & 54.98 & 63.3 & 46.59 & 75.8 & 60.0 \\
                                                  & QuaRot & - & - & - & - & - & - & - & - \\
                                                  & $\text{QuaRot}^{+}$ & 3411 & 0.06 & 0.12 & 0.0 & 49.72 & \textbf{23.98} & 27.99 & 16.98 \\
                                                  & $\text{QuaRot}^{+}$ + GBS & \textbf{34.97} & \textbf{13.84} & \textbf{14.81} & \textbf{12.87} & \textbf{52.17} & 23.72 & \textbf{38.89} & \textbf{26.05}\\
        \bottomrule
        
    \end{tabular}
    
    \label{tab:results_3bits}
\end{table}
\end{center}
\vspace{-2cm}
\subsection{Results}
\subsubsection{Gradual Binary Search performances}
Table \ref{tab:results_4bits} and Table \ref{tab:results_3bits} shows the results in 4 and 3 bits quantization on the perplexity and 6 benchmarks. In 4 bits our method GBS clearly outperforms previous methods for all models improving up to almost 6\% for LLaMA3-8B, 5\% on Qwen2.5 1.5B Instruct, 4\% on Mistral 7B and 3\% on Mistral 7B Instruct.

In 3 bits GBS made activation quantization possible with an increase of accuracy reaching 40\% for Mistral 7B (Table \ref{tab:results_3bits}). All other models have been greatly affected by GBS reducing the gap with 4 bits quantization. PPL is also significantly impacted by GBS reducing by a factor of 100 in the case of LLaMA3-8B. We now have a method reaching decent performances in 3 bits like with Mistral 7B Instruct which is only 10\% less than FP16 and reach 61.32\% accuracy on our benchmarks.

GBS appears to be highly effective in enhancing quantization performance, supporting our hypothesis that optimizing Perplexity via binary search is preferable to minimizing quantization error. We assumed a single minimum and a convex function, allowing us to leverage binary search while relying on the smoothness of CrossEntropy—an assumption that appears to hold true. Perplexity emerges as a strong objective for guiding our optimization, as it correlates well with improved benchmark performance (see Appendix \ref{sec:pplasobj} for more details).

We also evaluated GBS using two additional methods: SpinQuant \cite{liu_spinquant_2024} and DFRot \cite{xiang_dfrot_2024}, both relying on rotation matrices (see Appendix \ref{sec:more_results}). Our binary search approach, particularly when restricted to 3 bits, significantly enhances their performance, thus validating the broad effectiveness of our method

\begin{figure}[h]
    \centering
    \includegraphics[width=0.5\textwidth]{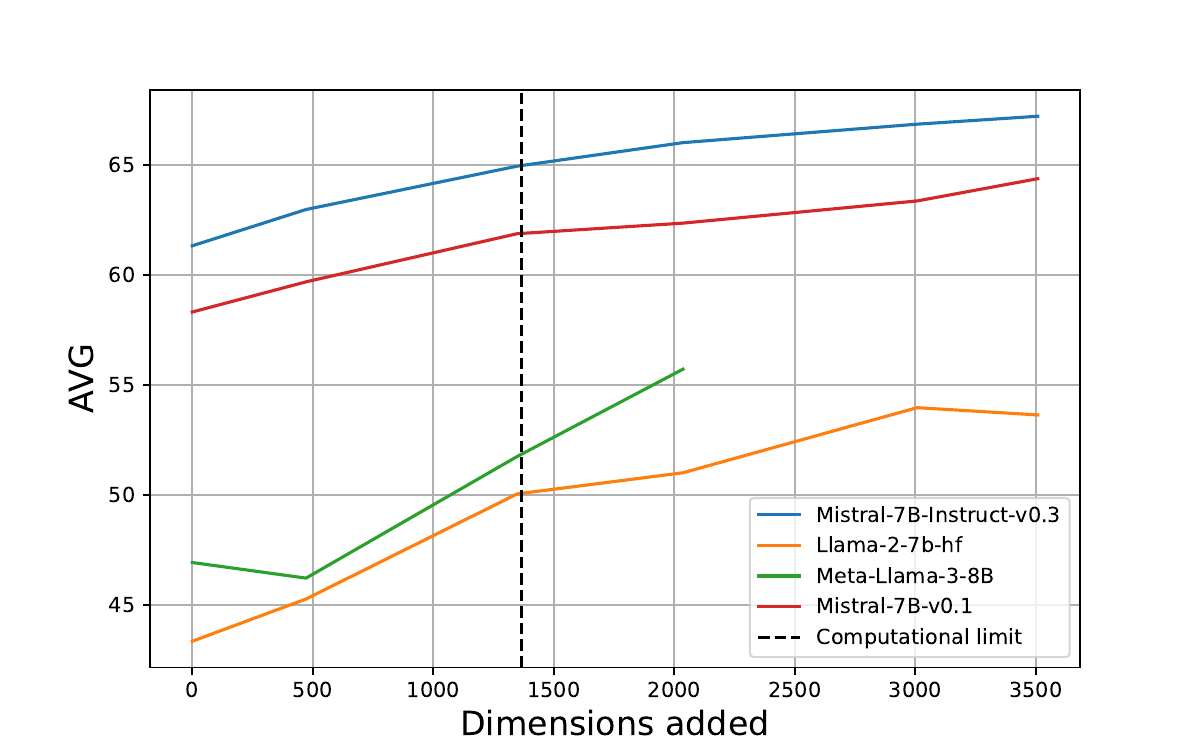}
        \caption{Effect of expanding dimensions on 6 benchmarks average (AVG) for different models in 3 bits WAKV quantization and the computational limit of Lemma \ref{lemme:3}. Due to memory constraints on GPU A100 we could not increase more than 2036 dimensions for LLaMA3-8B.}
    \label{fig:expanding_avg}
\end{figure}

\subsubsection{Matrix expansion effect}

To execute QuaRot on Qwen2.5 1.5B, as shown in Tables \ref{tab:results_4bits} and \ref{tab:results_3bits}, we needed to increase the embedding dimension (as detailed in Section \ref{sec:incr_dim}) to a value that can generate a Hadamard matrix. Initially, 1536 was not suitable since it is neither a power of 2 nor divisible by a known dimensions like 172, 156, or 140. We first chose to add 8 dimensions, reaching 1542, which can be managed by the Paley algorithm without excessive computational costs. This algorithm is also used for Qwen2.5 7B to generate a Hadamard matrix of dimension 3584 and expand MLP to 32 dimensions, enabling the Qwen architecture to work with QuaRot. This process is applicable to any architecture and any embedding dimension without increasing computational costs making this method very general.

We now study the impact of expanding dimensions on performance. Figure \ref{fig:expanding_avg} show the evolution of AVG with the number of dimensions added to our tokens and we clearly see the positive impact on performances. We can reach 68.95\% of accuracy for Mistral-7B Instruct but at a very high computational cost.

Another beneficial aspect of dimension expansion is seen in Group Local Rotation, introduced in QuaRot and explored in \cite{kim_lightrot_2025}. This technique involves decomposing a tensor into smaller sub-tensors and applying the same small power-of-2 Hadamard matrix to each of these sub-tensors. This approach leverages efficient Hadamard transforms (as introduced in Section \ref{sec:hadamard_matrix}) and significantly speeds up inference. Particularly for MLP layers that often operate in high-dimensional spaces, expanding dimensions can help identify a more suitable divisor, resulting in efficient power-of-2 sub-tensors.

\section{Conclusion}
In this work, we introduced an approach to optimize the quantization of LLMs using Gradual Binary Search and Hadamard matrices. Our method achieves efficient 3-bit quantization for weights, activations, and key-value caches, significantly improving model performance. We also theoretically demonstrated that Hadamard matrices are more effective than random rotation matrices in reducing extreme values in activations.

We also extended the use of rotation matrices to support non-power-of-2 embedding dimensions using the Paley algorithm and dimension expansion. This generalization allows our method to be applied to various architectures, including those with unique embedding dimensions. Experimental results on models from the Mistral library, LLaMA architecture, and Qwen show the effectiveness of our approach, outperforming existing methods.

Overall, our findings suggest that GBS and Hadamard matrices have great potential for advancing LLM quantization, making them more suitable for resource-constrained devices. Future work will explore mix computation and combining GBS with other methods.

\section{Limitations}

As explained in the previous part expanding dimensions has a big computational cost and it worsen with context length that is why we need to be aware of the expanding limit. One potential solution is to implement a Mixed Computation pipeline, where dimensions are only expanded in specific layers based on the presence of outliers, thereby substantially reducing computational overhead.

Another challenge arises with GBS, which involves computing perplexity at each step—a process that can be time-consuming for  PTQ methods, sometimes taking several days. To mitigate this, we could use less than 10\% of WikiText2 which might be excessive and unnecessary.

\newpage

\bibliography{neurips_2025}

\appendix

\section{Theoretical proofs}
\label{app:theoretical_results}
\subsection{Hadamard matrices}
\subsubsection{Proof}

\begin{proof}[Proof of Lemma \ref{lemme:1}]
We define a Hadamard matrix as a rotation matrix with values equal to 1 or -1 only. By definition, the equality $HH^T = I_n$ must be respected, but without normalization, we have $HH^T = nI_n$. If we decide to normalize this Hadamard matrix by a factor of $\frac{1}{\sqrt{n}}$, we obtain the identity by multiplying it by its transpose.
In the case of a vector $x = (c, \epsilon, ..., \epsilon)^T$ with $c >> \epsilon$, applying a Hadamard matrix to it amounts to multiplying the maximum absolute value by $\frac{1}{\sqrt{n}}$ since all the values of $H$ are either $\frac{1}{\sqrt{n}}$ or $- \frac{1}{\sqrt{n}}$, hence the desired result.
\end{proof}

\begin{proof}[Proof of Lemma \ref{lemme:2}]
Let $Q$ be a rotation matrix drawn on the unit sphere $\mathcal{S}^{n-1}$. We assume that the problem is in high dimension, which allows us to approximate the distribution of the elements of the matrix $Q$:
$$
Q_{ij} \sim \mathcal{N}\left(0,\frac{1}{n} \right)
$$
This theorem is a classic result of high-dimensional probability theory \cite{vershynin_high-dimensional_2018}. It is this result that allows the entire demonstration, because it is from this approximation that we can use the fundamental properties of the extreme values of a normal law.
Indeed, for all $i,j \leq n,\,  Z_{ij} \sim \mathcal{N}(0,1)$, then $Q_{ij} = \frac{Z_{ij}}{\sqrt{n}}$. We can show \cite{de_haan_extreme_2006} that
\begin{equation}
    \mathbb{E}\left[\max_{1\leq i,j \leq n}|Z_{ij}|\right] = \sqrt{2\log n}
\end{equation}
Therefore,
\begin{equation}
    \mathbb{E}\left[\max_{1\leq i,j \leq n}|Q_{ij}|\right] = \sqrt{\frac{2\log n}{n}}
\end{equation}
Using Talagrand's inequality for a Lipschitz function ($Qx$ is indeed a Lipschitz function) we have:
\begin{equation}
    \mathbb{P}\left(\left|\max_{1\leq i,j \leq n}|Q_{ij}|-\sqrt{\frac{2\log n}{n}}\right| > \epsilon \right) \leq 2e^{-Cn\epsilon^2}
\end{equation}
With $C > 0$. Thus, for sufficiently large $n$, we have a very high probability of having:
\begin{equation}
    \max_{1\leq i,j \leq n}|Q_{ij}| = \sqrt{\frac{2\log n}{n}}
\end{equation}
We now use this result when applying $Q$ to a vector $x = (c, \epsilon, ..., \epsilon)^T$ with $c >> \epsilon$
$$\max_{1 \leq i \leq n}|(Qx)_i| = c\max_{1 \leq i,j \leq n}|Q_{ij}|= c\sqrt{\frac{2\log n}{n}}$$
\end{proof}
Finally, using Lemmas \ref{lemme:1} and \ref{lemme:2}, we show that for sufficiently large $n$
$$\max_{1 \leq i \leq n}|(Hx)_i| \leq \max_{1 \leq i \leq n}|(Qx)_i|$$
We even show that we cannot do better than the Hadamard matrix to redistribute the energy of a matrix.

\begin{proof}[Proof of Theorem \ref{thm:HdmOpt}]
Let $Q \in \mathbb{R}^{n\times n}$ be an orthogonal matrix. By definition, for any column $i$ of this matrix:
$$
||Q_i||_2 = 1 \Rightarrow \sum_{j}|Q_{i,j}|^2 = 1
$$
However,
\begin{gather}
\sum_{j}\max_{p,k}|Q_{p,k}|^2 \geq \sum_{j}|Q_{i,j}|^2 \\
\Rightarrow \sum_{j}\max_{p,k}|Q_{p,k}|^2 \geq 1 \\
\Rightarrow n * \max_{p,k}|Q_{p,k}|^2 \geq 1 \\
\Rightarrow \max_{p,k}|Q_{p,k}| \geq \frac{1}{\sqrt{n}}
\end{gather}
We have just shown that an orthogonal matrix cannot reduce an outlier by more than a factor of $\frac{1}{\sqrt{n}}$, and as seen with Lemma \ref{lemme:1}, a Hadamard matrix can reach this bound, it is therefore optimal.
\end{proof}

\subsubsection{Experimental Verifications}
\begin{figure}[h]
    \centering
    \includegraphics[width=0.8\linewidth]{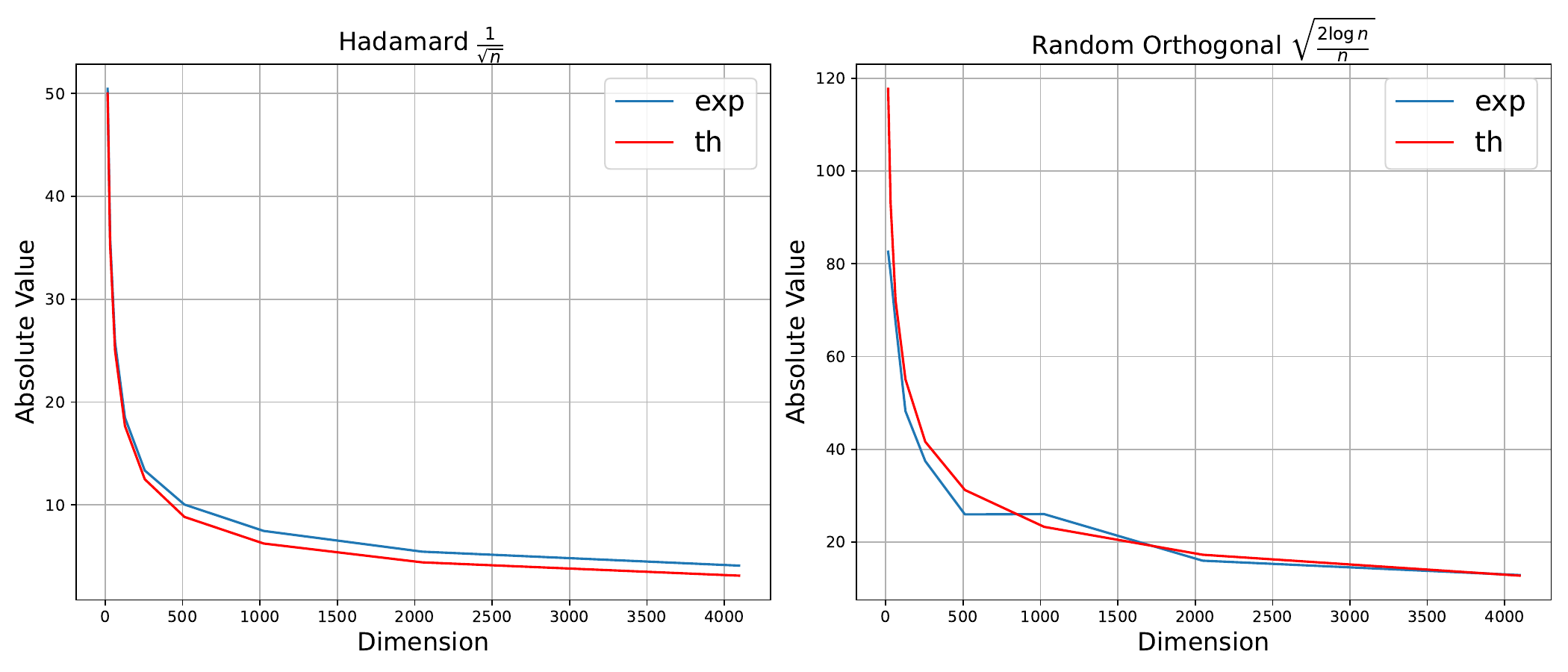}
    \caption{Maximum absolute value as a function of dimension for a randomly drawn rotation matrix and a Hadamard matrix applied to a vector containing a peak at 200 obtained experimentally (blue) and theoretically (red)}
    \label{fig:AbsMax}
\end{figure}

To verify the previous theoretical results, we set up a simple experiment where we apply a randomly drawn rotation matrix and a Hadamard matrix to a vector of dimension $n$ following a standard normal distribution with a standard deviation of 0.1 to which we add a peak at 200.
We plot the curve of the maximum absolute value after applying our matrices as a function of the dimension $n$ as well as the theoretical curve in Figue \ref{fig:AbsMax}.

It clearly shows that the theoretical and experimental curves follow each other perfectly, which seems to confirm the previously demonstrated theorems. Hadamard matrices are therefore theoretically and experimentally the most suitable matrices for reducing the impact of an outlier in a vector.

\subsection{Increasing dimensions}
\begin{proof}[Proof of Lemma \ref{lemme:3}]
We define BitOps as the function that compute the number of operations for a matrix multiplication:
$\text{BitOps}(AB) = mn^2pb^2$. And we want to find a condition that ensure 
\begin{align}
    \text{BitOps}(A'B') &\leq \text{BitOps}(AB) \\
    \Rightarrow\quad m(n+d)^2pb'^2 &\leq mn^2pb^2 \\
    \Rightarrow\quad (n+d)^2b'^2 &\leq n^2b^2 \\
    \Rightarrow\quad (n+d)b' &\leq nb \\
    \Rightarrow\quad nb'+db' &\leq nb \\
    \Rightarrow\quad d &\leq \frac{n(b-b')}{b'}
\end{align}

\end{proof}

\section{Paley Algorithm}
\begin{algorithm}[h]
\caption{Hadamard Matrix Construction using the Paley Method}
\label{alg:paley}
\begin{algorithmic}[1]
\Require A prime number \( p \)  
\Ensure A Hadamard matrix \( H \) of order \( p + 1 \) 
\State \( n \gets p + 1 \)  \Comment{Determine the order of the matrix}
\State Initialize \( H \) as a \( n \times n \) matrix with all entries set to 1  \Comment{Start with a matrix of all ones}
\For{\( i \gets 1 \) to \( n-1 \)}  \Comment{Loop over rows (except the first row)}
    \State \( H[i, 0] \gets -1 \)  \Comment{Set the first column entry to -1 for current row}
    \State \( H[0, i] \gets -1 \)  \Comment{Set the first row entry to -1 for current column}
    \For{\( j \gets 1 \) to \( n-1 \)}  \Comment{Loop over columns (except the first column)}
        \If{\( i = j \)}  
            \State \( H[i, j] \gets -1 \)  \Comment{Set diagonal entries to -1}
        \Else 
            \State \( H[i, j] \gets \text{legendre\_symbol}((i-1) - (j-1), p) \)  
        \EndIf
    \EndFor
\EndFor
\State \Return \( H \)  \Comment{Return the constructed Hadamard matrix}
\end{algorithmic}
\end{algorithm}

\section{Gradual Binary Search process}
\label{app:GBS_process}

\begin{figure}[ht]
    \centering
    \subfigure[GBS starting in 4 bits]{
        \includegraphics[width=0.48\textwidth]{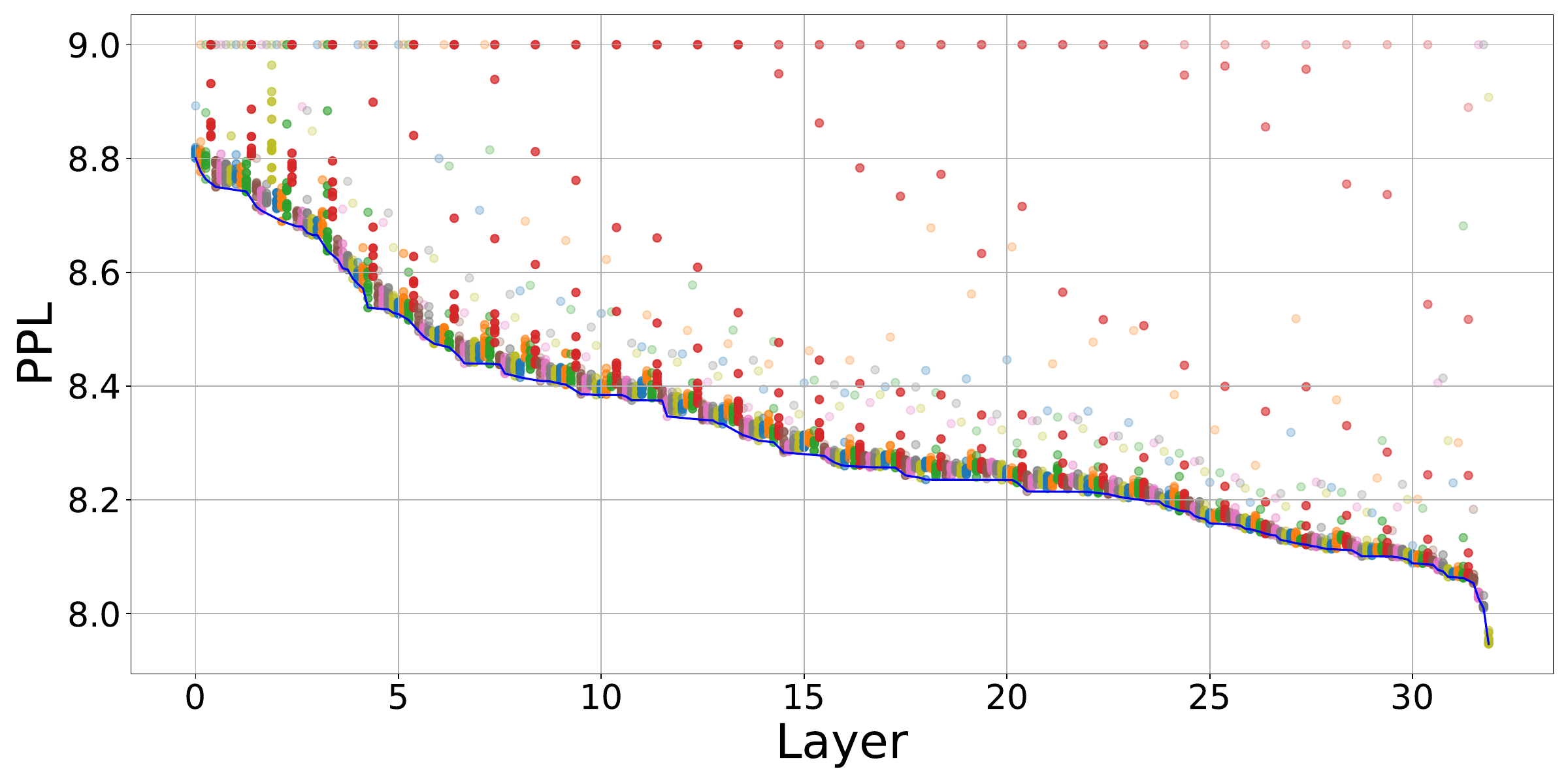}
        \label{fig:pplvslayer_16bits}
    }
    \subfigure[GBS starting in FP16]{
        \includegraphics[width=0.48\textwidth]{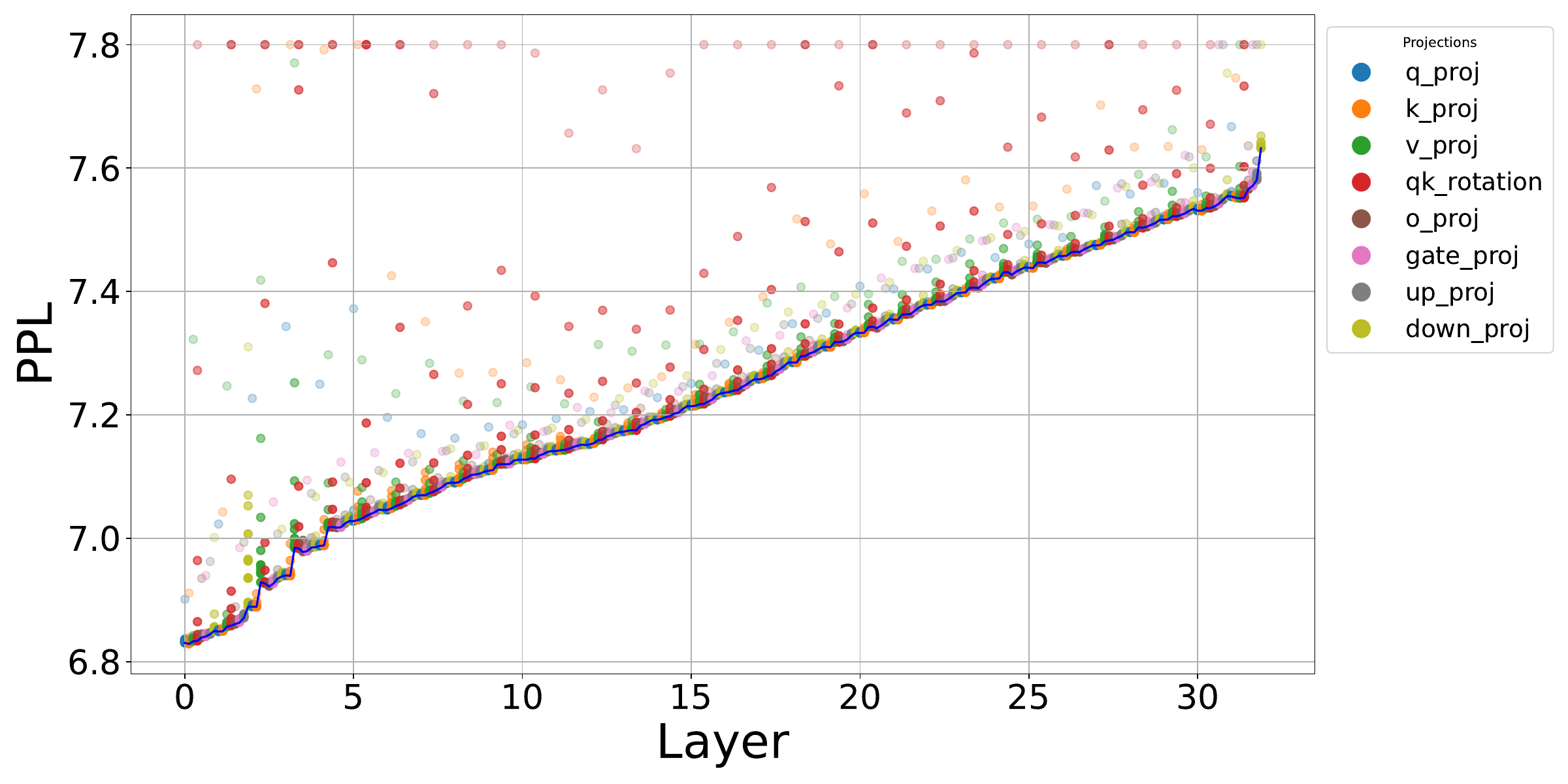}
        \label{fig:pplvslayer_4bits}
    }
    \caption{PPL vs Layer during Gradual Binary Search on 10\% of Train WikiText2 for a LLaMA3-8B in 4-bit quantization and rotated with QuaRot. For better visualization we set a maximum PPL to 9. Points opacity represents the clipping ratio, the value is closer to 0 as transparency increases}
    \label{fig:pplvslayer}
\end{figure}

In this analysis, we examine the evolution of clipping ratios through GBS to better understand the dynamics of these parameters in LLMs. Figure \ref{fig:pplvslayer} illustrates the perplexity (PPL) evolution during the optimization of a LLaMA3-8B model quantized to 4 bits. The graph displays the various tested values for each projection, optimized under two different configurations: starting the model in FP16 (blue line) and initiating the process in 4 bits. It is clear that starting in FP16 yields a better PPL on the training set of WikiText2, achieving 7.62, compared to starting in 4 bits, which results in a PPL of 7.94. On the test set we have the same dynamic with a PPL of 7.4 starting in FP16 and 7.69 starting in 4 bits.

\begin{figure}[ht]
    \centering
    \subfigure[GBS starting in FP16]{
        \includegraphics[width=0.48\textwidth]{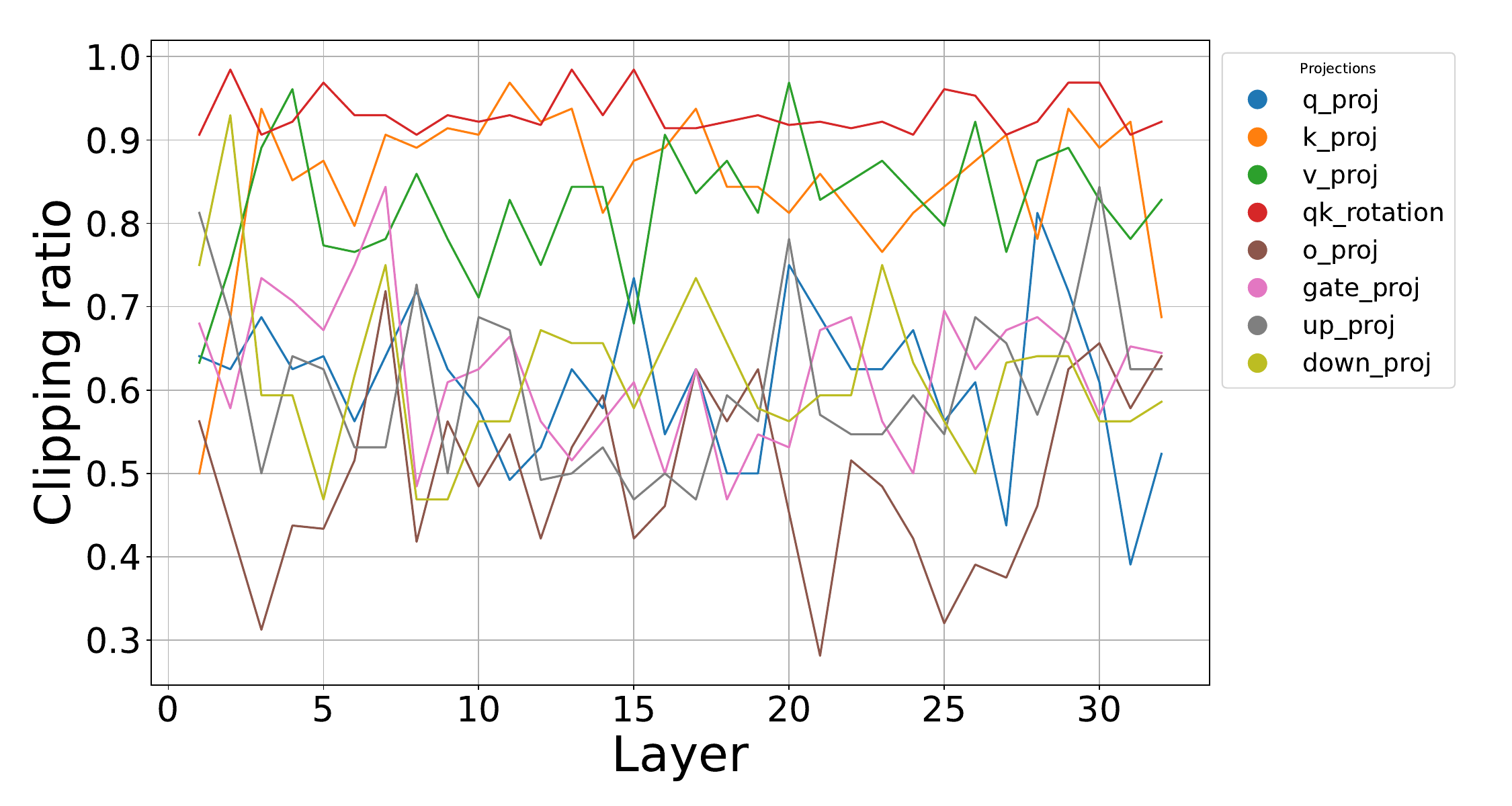}
        \label{fig:crvslayer_FP16}
    }
    \subfigure[GBS starting in 4 bits]{
        \includegraphics[width=0.48\textwidth]{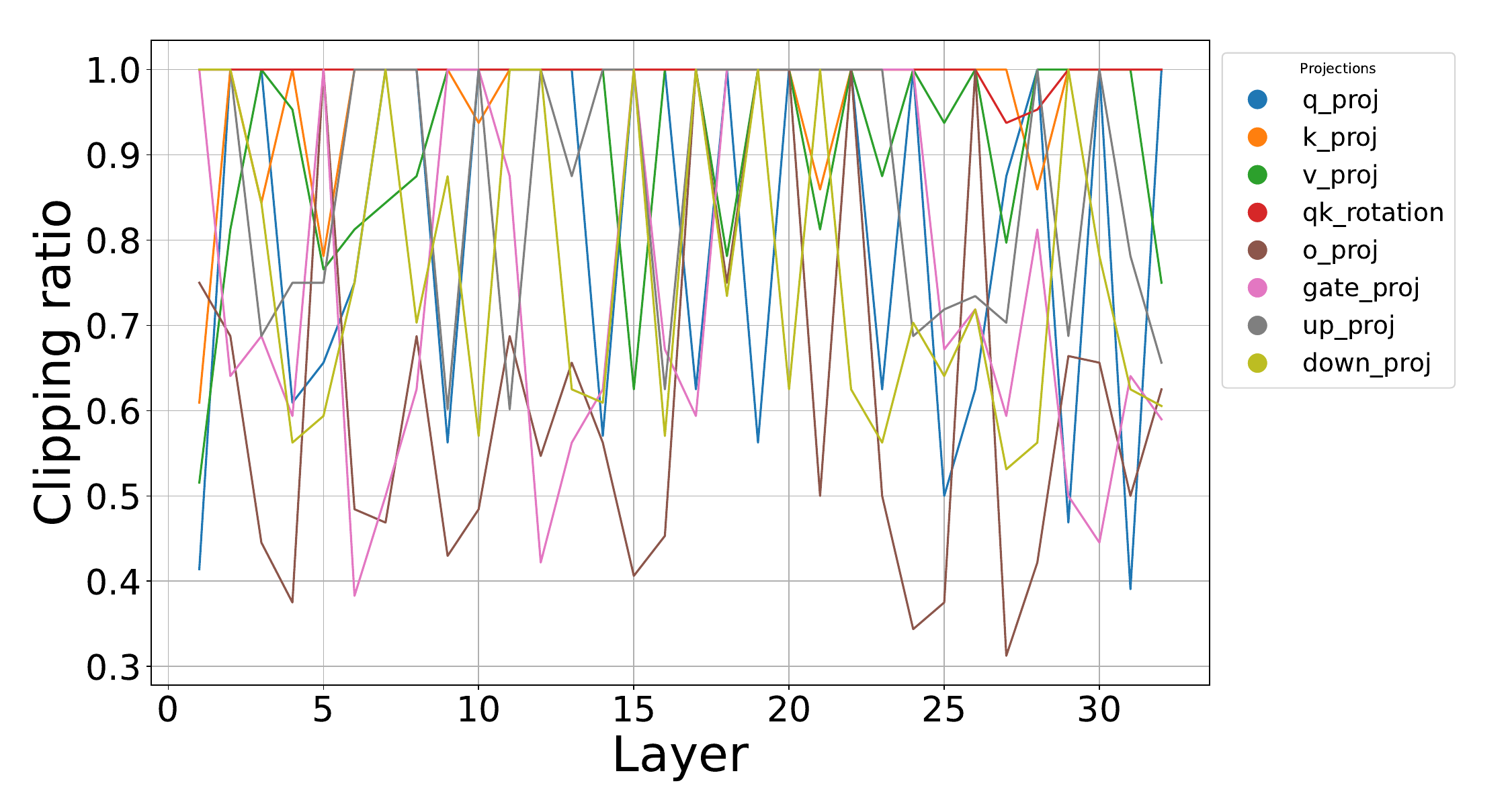}
        \label{fig:crvslayer_4bits}
    }
    \caption{Final configurations obtained with GBS started in 4 bits and in FP16 for a LLaMA3-8B}
    \label{fig:crvslayer}
\end{figure}

Figure \ref{fig:crvslayer} illustrates the final configuration achieved by GBS for the same architecture, starting the process in both FP16 and 4-bit precision. It is clear that initiating in 4 bits results in a significantly more unstable configuration compared to starting in FP16. Many values remain at 1, and there is a high variance, indicating that the algorithm struggles to find a stable configuration when the entire model is in 4 bits. It also appears to have difficulty understanding the impact of small changes in the clipping ratio.

In contrast, starting in FP16 results in a stable configuration for every projection, with distinct dynamics. For instance, the $\text{qk\_rotation}$ projection exhibits minimal changes in the clipping ratio, with most layers close to 1. Conversely, the $\text{o\_proj}$ projection has values below 0.3, suggesting that clipping to 30\% of the maximum value can enhance performance. This figure underscores the importance of GBS in improving the model's quality by identifying the optimal clipping configuration, which is clearly not only ones.

\section{Perplexity as objective}
\label{sec:pplasobj}

\begin{figure}[h]
    \centering
    \includegraphics[width=0.4\textwidth]{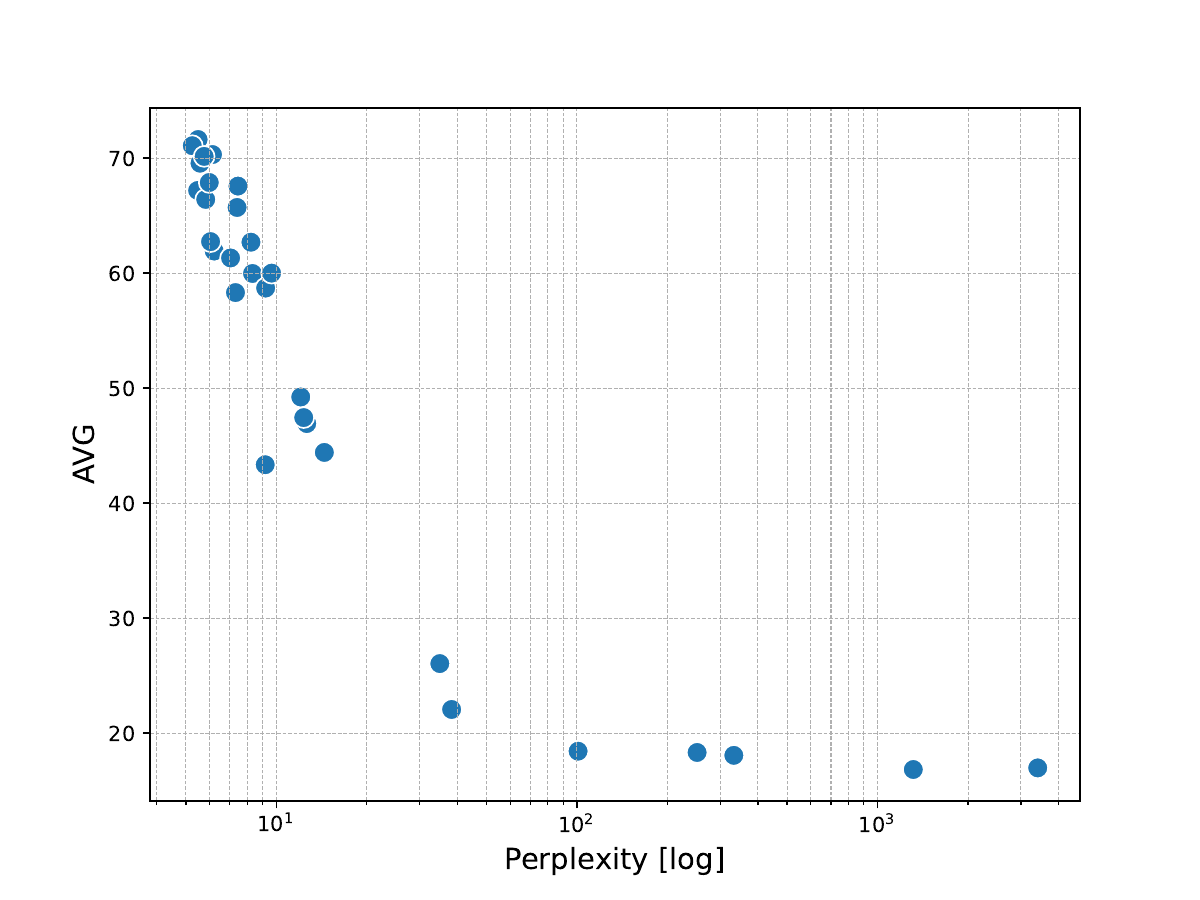}
    \caption{AVG over Perplexity for all results obtained in Tab \ref{tab:results_3bits} and \ref{tab:results_4bits}}
    \label{fig:avgvsppl}
\end{figure}

Perplexity is the central part of our optimization, it drives our search and it is supposed to reach a configuration which will performs better than all others with a bigger PPL. In figure \ref{fig:avgvsppl} we can see how the average value on 6 benchmarks evolves with the perplexity. It clearly appears that a smaller PPL usually represents a better AVG.  

\section{More results}
\label{sec:more_results}
\subsection{SpinQuant}
\begin{center}
    
\begin{table}[h]
    \footnotesize
    \centering
    \renewcommand{\arraystretch}{1.05}
    \setlength{\tabcolsep}{3pt}
    \caption{Results in 4 bits WAKV quantization on perplexity (PPL), PIQA, hellaswag (HS), arc-easy (ARC-E), arc-challenge (ARC-C), winogrande (WINO) and lambada, we also compute the average value (AVG) which represents a \% of success. We compare our method, GBS, with SpinQuant  and clearly observe that GBS outperforms SpinQuant across almost all computed metrics. }
    \begin{tabular}{cccccccccc}
        \toprule
        \textbf{Model} & \textbf{Method} & \textbf{PPL}$\downarrow$ & \textbf{PIQA} & \textbf{HS} & \textbf{ARC-E} & \textbf{ARC-C} & \textbf{Wino} & \textbf{Lambada} & \textbf{AVG}$\uparrow$ \\
        \hline
        \multirow{3}{*}{Mistral 7B Inst v0.3} & FP16 & 5.49  & 71.27 & 74.6 & 67.94 & 74.27 & 58.96 & 82.66 & 71.62 \\
                                                  & SpinQuant & 5.88  & 69& 72.66 & 65.34 & 69.93 & 55.12 & 78.41 & 68.41 \\
                                                  & SpinQuant + GBS & \textbf{5.83 }& \textbf{69.01} & \textbf{72.66} & \textbf{65.36} & \textbf{72.14} & \textbf{56.91} & \textbf{79.5} & \textbf{69.26}\\
        \hline
        \multirow{3}{*}{Mistral 7B v0.1} & FP16 & 5.25 & 72.49 & 75.59 & 69.4 & 73.95 & 54.86 & 80.18 & 71.08 \\
                                                  
                                                  & SpinQuant & 5.71 & 69.55 & \textbf{73.45} & 65.65 & 69.38 & \textbf{48.98} & \textbf{76.98} & 67.33 \\
                                                  & SpinQuant + GBS & \textbf{5.62} & \textbf{70.02} & 73.37 & \textbf{66.66} & \textbf{71.19} & 48.29 & 76.47 & \textbf{67.67} \\
        \hline
        \multirow{3}{*}{Llama2 7B} & FP16 & 5.47 & 71.08 & 73.9 & 68.25 & 68.98 & 46.33 & 74.58 & 67.19 \\
                                                  
                                                  & SpinQuant & 6.57 & 61.73 & 68.46 & 55.0 & 63.46 & 40.61 & 67.63 & 59.48 \\
                                                  & SpinQuant + GBS & \textbf{6.15} & \textbf{63.24} & \textbf{69.01} & \textbf{57.46} & \textbf{65.04} & \textbf{41.04} & \textbf{68.43} & \textbf{60.7} \\
        \hline
        \multirow{3}{*}{Llama3 8B} & FP16 & 6.13 & 72.62 & 76.01 & 69.22 & 72.93 & 53.41 & 77.69 & 70.3 \\
                                                  
                                                  & SpinQuant & 7.97 & \textbf{65.11} & \textbf{69.01} & \textbf{61.21} & 66.61 & 45.22 & 72.52 & 63.28 \\
                                                  & SpinQuant + GBS & \textbf{7.69 }& 63.84 & 67.84 & 59.83 & \textbf{70.4 }& \textbf{47.35} & \textbf{73.36} & \textbf{63.77} \\
        \bottomrule
        
    \end{tabular}
    
    \label{tab:results_4bits_spinquant}
\end{table}
\end{center}

\begin{center}
    
\begin{table}[h]
    \footnotesize
    \renewcommand{\arraystretch}{1.05}
    \setlength{\tabcolsep}{3pt}
    \caption{Results in 3 bits WAKV quantization on perplexity (PPL), PIQA, hellaswag (HS), arc-easy (ARC-E), arc-challenge (ARC-C), winogrande (WINO) and lambada, we also compute the average value (AVG) which represents a \% of success. We compare our method, GBS, with SpinQuant  and clearly observe that GBS outperforms SpinQuant across almost all computed metrics. }
    \begin{tabular}{cccccccccc}
        \toprule
        \textbf{Model} & \textbf{Method} & \textbf{PPL}$\downarrow$ & \textbf{PIQA} & \textbf{HS} & \textbf{ARC-E} & \textbf{ARC-C} & \textbf{Wino} & \textbf{Lambada} & \textbf{AVG}$\uparrow$ \\
        \hline
        \multirow{3}{*}{Mistral 7B Inst v0.3} & FP16 & 5.49  & 71.27 & 74.6 & 67.94 & 74.27 & 58.96 & 82.66 & 71.62 \\
                                                  & SpinQuant & 14.31 & 20.39 & 28.39 & 12.38 & 49.72 & 24.57 & 40.57 & 29.34 \\
                                                  & SpinQuant + GBS & \textbf{8.11 }& \textbf{52.93} & \textbf{58.99} & \textbf{46.87} & \textbf{60.06} & \textbf{40.78} & \textbf{66.67} & \textbf{54.38}\\
        \hline
        \multirow{3}{*}{Mistral 7B v0.1} & FP16 & 5.25 & 72.49 & 75.59 & 69.4 & 73.95 & 54.86 & 80.18 & 71.08 \\
                                                  
                                                  & SpinQuant & 18.88 & 16.58 & 21.97 & 11.2 & 51.3 & 24.23 & 37.67 & 27.16 \\
                                                  & SpinQuant + GBS & \textbf{7.98} & \textbf{52.81} & \textbf{60.59} & \textbf{45.04} & \textbf{59.27} & \textbf{35.67} & \textbf{63.68} & \textbf{52.84} \\
        \hline
        \multirow{3}{*}{Llama2 7B} & FP16 & 5.47 & 71.08 & 73.9 & 68.25 & 68.98 & 46.33 & 74.58 & 67.19 \\
                                                  
                                                  & SpinQuant & 425.2 & 0.24 & 0.45 & 0.04 & \textbf{51.46} & 28.33 & 27.57 & 18.02 \\
                                                  & SpinQuant + GBS & \textbf{15.65} & \textbf{20.77} & \textbf{26.51} & \textbf{15.04} & 50.67 & \textbf{26.19} & \textbf{42.93} & \textbf{30.35} \\
        \hline
        \multirow{3}{*}{Llama3 8B} & FP16 & 6.13 & 72.62 & 76.01 & 69.22 & 72.93 & 53.41 & 77.69 & 70.3 \\
                                                  
                                                  & SpinQuant & 316.6 & 2.71 & 3.1 & 2.31 & 50.51 & 22.35 & 29.0 & 18.33\\
                                                  & SpinQuant + GBS & \textbf{20.26}&  \textbf{21.42} & \textbf{23.95} & \textbf{18.9 }& \textbf{54.38} & \textbf{26.96} & \textbf{43.1 }& \textbf{31.45} \\
        \bottomrule
        
    \end{tabular}
    
    \label{tab:results_4bits_spinquant}
\end{table}
\end{center}

\newpage
\subsection{DFRot}

\begin{center}
    
\begin{table}[h]
    \footnotesize
    \renewcommand{\arraystretch}{1.05}
    \setlength{\tabcolsep}{3pt}
    \caption{Results in 4 bits WAKV quantization on perplexity (PPL), PIQA, hellaswag (HS), arc-easy (ARC-E), arc-challenge (ARC-C), winogrande (WINO) and lambada, we also compute the average value (AVG) which represents a \% of success. We compare our method, GBS, with DFRot and clearly observe that GBS outperforms DFRot across all computed metrics. }
    \begin{tabular}{cccccccccc}
        \toprule
        \textbf{Model} & \textbf{Method} & \textbf{PPL}$\downarrow$ & \textbf{PIQA} & \textbf{HS} & \textbf{ARC-E} & \textbf{ARC-C} & \textbf{Wino} & \textbf{Lambada} & \textbf{AVG}$\uparrow$ \\
        \hline
        \multirow{3}{*}{Mistral 7B Inst v0.3} & FP16 & 5.49  & 71.27 & 74.6 & 67.94 & 74.27 & 58.96 & 82.66 & 71.62 \\
                                                  & DFRot & 5.94 &68.11 & 79.43 & 68.5 & 72.42 & 64.58 & \textbf{80.3 }& 72.22 \\
                                                  & DFRot + GBS & \textbf{5.81 }& \textbf{71.35} & \textbf{81.23} & \textbf{69.3 }& \textbf{73.18} & \textbf{65.42} & 80.13 & \textbf{73.43} \\
        \hline
        \multirow{3}{*}{Mistral 7B v0.1} & FP16 & 5.25 & 72.49 & 75.59 & 69.4 & 73.95 & 54.86 & 80.18 & 71.08 \\
                                                  
                                                  & DFRot & 5.75 &\textbf{71.03} & 78.94 & 69.83 & 74.03 & 65.63 & 78.28 & 72.96 \\
                                                  & DFRot + GBS & \textbf{5.62 }& 70.88 & \textbf{80.9 }& \textbf{70.93} & \textbf{75.0 }& \textbf{66.85} & \textbf{78.85} & \textbf{73.9 }\\
        \hline
        \multirow{3}{*}{Llama2 7B} & FP16 & 5.47 & 71.08 & 73.9 & 68.25 & 68.98 & 46.33 & 74.58 & 67.19 \\
                                                  
                                                  & DFRot & 6.23 & 65.04 & 76.66 & 65.75 & 69.47 & 62.02 & 72.61 & 68.59\\
                                                  & DFRot + GBS & \textbf{6.05 }& \textbf{65.67} & \textbf{77.75} & \textbf{66.41} & \textbf{69.63} & \textbf{63.19} & \textbf{72.74} & \textbf{69.23}\\
        \hline
        \multirow{3}{*}{Llama3 8B} & FP16 & 6.13 & 72.62 & 76.01 & 69.22 & 72.93 & 53.41 & 77.69 & 70.3 \\
                                                  
                                                  & DFRot & 7.95 & 68.11 & 76.01 & 64.92 & 68.5 & 61.34 & 74.17 & 68.84 \\
                                                  & DFRot + GBS & \textbf{7.56 }& \textbf{72.53} & \textbf{76.82} & \textbf{66.35} & \textbf{69.53} & \textbf{63.17} & \textbf{75.0 }& \textbf{70.57} \\
        \bottomrule
        
    \end{tabular}
    
    \label{tab:results_4bits_dfrot}
\end{table}
\end{center}

\begin{center}
    
\begin{table}[h]
    \small
    \renewcommand{\arraystretch}{1.05}
    \setlength{\tabcolsep}{3pt}
    \centering
    \caption{Results in 3 bits WAKV quantization on perplexity (PPL), PIQA, hellaswag (HS), arc-easy (ARC-E), arc-challenge (ARC-C), winogrande (WINO) and lambada, we also compute the average value (AVG) which represents a \% of success. We compare our method, GBS, with DFRot  and clearly observe that GBS outperforms DFRot across all computed metrics. }
    \begin{tabular}{cccccccccc}
        \toprule
        \textbf{Model} & \textbf{Method} & \textbf{PPL}$\downarrow$ & \textbf{PIQA} & \textbf{HS} & \textbf{ARC-E} & \textbf{ARC-C} & \textbf{Wino} & \textbf{Lambada} & \textbf{AVG}$\uparrow$ \\
        \hline
        \multirow{3}{*}{Mistral 7B Inst v0.3} & FP16 & 5.49  & 71.27 & 74.6 & 67.94 & 74.27 & 58.96 & 82.66 & 71.62 \\
                                                  & DFRot & 11.26 & 53.28 & 67.57 & 35.6 & 40.33 & 30.88 & 57.57 & 47.54\\
                                                  & DFRot + GBS & \textbf{7.58 }& \textbf{63.22} & \textbf{75.35} & \textbf{59.32} & \textbf{65.19} & \textbf{53.46} & \textbf{71.73} & \textbf{64.71} \\
        \hline
        \multirow{3}{*}{Mistral 7B v0.1} & FP16 & 5.25 & 72.49 & 75.59 & 69.4 & 73.95 & 54.86 & 80.18 & 71.08 \\
                                                  
                                                  & DFRot & 13.63 & 55.01 & 65.45 & 27.89 & 32.52 & 23.25 & 50.63 & 42.46\\
                                                  & DFRot + GBS & \textbf{7.64 }& \textbf{62.83} & \textbf{73.72} & \textbf{56.99} & \textbf{64.33} & \textbf{49.64} & \textbf{68.12} & \textbf{62.6}\\
        \hline
        \multirow{3}{*}{Llama2 7B} & FP16 & 5.47 & 71.08 & 73.9 & 68.25 & 68.98 & 46.33 & 74.58 & 67.19 \\
                                                  
                                                  & DFRot & 26.64 & 49.96 & 58.81 & 13.19 & 14.81 & 11.57 & 39.14 & 31.25\\
                                                  & DFRot + GBS & \textbf{10.96} & \textbf{56.12} & \textbf{66.81} & \textbf{34.43} & \textbf{40.4 }& \textbf{28.45} & \textbf{55.01} & \textbf{46.87} \\
        \hline
        \multirow{3}{*}{Llama3 8B} & FP16 & 6.13 & 72.62 & 76.01 & 69.22 & 72.93 & 53.41 & 77.69 & 70.3 \\
                                                  
                                                  & DFRot & 140.78 & 52.41 & 54.62 & 2.82 & 3.38 & 2.27 & 31.85 & 24.56\\
                                                  & DFRot + GBS & \textbf{22.14} & \textbf{54.85} & \textbf{61.92} & \textbf{25.53} & \textbf{29.58} & \textbf{21.48} & \textbf{48.67} & \textbf{40.34}\\
        \bottomrule
        
    \end{tabular}
    
    \label{tab:results_3bits_dfrot}
\end{table}
\end{center}

\end{document}